\pdfoutput=1
\documentclass[11pt]{article}

\usepackage{CJKutf8}
\usepackage{acl}

\usepackage{times}
\usepackage{latexsym}
\usepackage{booktabs}
\usepackage{graphicx}
\usepackage{pmboxdraw}
\newcommand\boxRight{\textSFii\pmboxdrawuni{2574}}
\usepackage{enumitem}
\usepackage[T1]{fontenc}

\usepackage[utf8]{inputenc}

\usepackage{microtype}

\title{Knowledge Based Template Machine Translation In Low-Resource Setting}

\author{Zilu Tang  \\
  Boston University \\
  \texttt{zilutang@bu.edu} \\\And
  Derry Tanti Wijaya \\
  Boston University \\
  \texttt{wijaya@bu.edu} \\}

\begin{document}
\maketitle
\begin{abstract}
Incorporating tagging into neural machine translation (NMT) systems has shown promising results in helping translate rare words such as named entities (NE). However, translating NE in low-resource setting remains a challenge. In this work, we investigate the effect of using tags and NE hypernyms from knowledge graphs (KGs) in parallel corpus in different levels of resource conditions. We find the tag-and-copy mechanism (tag the NEs in the source sentence and copy them to the target sentence) improves translation in high-resource settings only. Introducing copying also results in polarizing effects in translating different parts-of-speech (POS). Interestingly, we find that copy accuracy for hypernyms is consistently higher than that of entities. As a way of avoiding "hard" copying and utilizing hypernym in bootstrapping rare entities, we introduced a "soft" tagging mechanism and found consistent improvement in high and low-resource settings.
\end{abstract}

\section{Introduction}

NMT methods usually require significant training data. For low-resource languages, NMT models generally do not work as well, especially when translating NEs. With low occurrences and large variations, NEs often remain unseen until inference time. In this paper, we investigate the usefulness of using template tagging methods and hypernyms to generalize NMT under low-resource settings. 
\vspace{-.5em}
\paragraph{Template Machine Translation}
Template NMT usually involves tagging the input sentences such that the templates simplify the task for the model during translation. While the idea is not new in statistical MT \citep{pal2010handling}, one of the first works in NMT addressing rare words in translation uses multiple numbered unknown (\textit{unks}) tokens to link up source and target sentences \citep{luong2015addressing}. With the introduction of such copy mechanism, models only need to copy (instead of translate) the unknown token from source to target sentence, and (if needed) perform post-processing to replace the copied-over tags. \citet{li2018neural} replaces named entities with their type symbols (i.e. LOC, ORG) on both source and target side, and trains a character-level sequence to sequence model for NE translation. \citet{crego2016systran} and \citet{wang2017sogou} use similar tagging mechanism, with the latter using a dictionary to translate tagged NE. \citet{wang2019merging} and  \citet{li2018named} use a few tagging methods from code-switching, boundary tags (i.e. \textit{<ORG>}, \textit{<}$\textbackslash$\textit{ORG>}), to extra embedding to tag NE on source and target side. Others have explored encouraging copying through constrained decoding (\citealp{hokamp2017lexically}, \citealp{post2018fast}), and modifying architecture or input (\citealp{gu2018search}, \citealp{pham2018towards} \citealp{dinu2019training}).
\vspace{-.5em}
\paragraph{Knowledge Augmented Translation}
In addition to tagging boundaries of NEs from previous section, a few methods also use POS and other linguistic features to improve NMT (\citealp{sennrich2016linguistic}, \citealp{modrzejewski2020incorporating}, \citealp{hamalainen2019template}). \citet{salam2017improve} uses hypernyms in a statistical MT system for low-resource translation. Meanwhile, many have used KGs to improve NMT systems. Some use KGs for data augmentation (\citealp{zhao2021knowledge}), while others combine NMT with knowledge graph embedding to improve translation quality (\citealp{lu2018exploiting}, \citealp{zhao2020knowledge}, \citet{moussallem2019augmenting}. 

While our goal resembles similar efforts in template machine translation, we extend the tag types to a much wider range using hypernyms obtained through KGs. In addition, we perform extensive analysis to understand the pros and cons of copy mechanism under different resource conditions. Our paper provides 3 key insights:

\begin{itemize}[noitemsep,topsep=0pt]
    \item Copy mechanism improves translation only in high-resource setting.
    \item Copy models translate syntactic POS better and semantic POS worse, yielding translation with similar sentence structures as the source.
    \item Appending hypernyms to NEs can improve translation accuracy in low-resource settings.
\end{itemize}

\section{Methods}

We first use statistical word alignment (WA) (FastAlign, \citealp{dyer2013simple}) to build a word translation table. We then use DBpedia Spotlight entity linking (EL) system \cite{isem2011mendesetal}\footnote{https://www.dbpedia-spotlight.org/}) to find NEs within sentences that connects to English DBpedia\footnote{https://www.dbpedia.org/}, as well as the translation of the NEs on target side through WA. We substitute the NEs with corresponding templates. After model translation, we remove the tags\footnote{Our soft tagging approach, \textbf{HypA}, does not contain explicit tags and requires no removal post translation}, either keep the translation already in the tag or use the word translation table to translate copied entities. Our system is modular and code can be found in our repo\footnote{Anonymized. Our code is included in a zip file as software component in the submission}.
\vspace{-1em}
\paragraph{Tagging Methods}
We use the following templates in our experiments (illustrated in Table~\ref{table:tag-list}): \textbf{Tag} and  \textbf{Trans} are similar to previous works shown to improve translation adequacy (\citealp{wang2019merging}, \citealp{li2018named}). We also experiment with adding semantic information in tags by appending entity's hypernym provided by DBpedia. Since hypernyms are more generalized with higher term frequency, we expect translation models to use them as context when translating sentences in addition to using them to copy. \textbf{Add} \textit{adds} hypernym after entity tag, \textbf{TransA} adds hypernym \textit{and} translation, while \textbf{TransR} \textit{replaces} original entity with hypernym and adds translation. For target sentences, we replace the NE translations (obtained by WA) with the same templates as the source sentences.

In addition to enforcing a "hard" copying mechanism using tagging templates, we also include a "soft" signal by adding the hypernym after the entity (\textbf{HypA}) without \textit{explicit} tags. On the target side, we append the translated hypernym if possible (from word translation table) otherwise we use the source language hypernym. Without an explicit signal for copying, we expect the model to rely on the hypernyms as context when translating NEs.

In our experiments, we ensure the same NEs are tagged across templates, with about 25\% of all sentences tagged in each dataset (Appx. Table~\ref{table:tag-stat}).

\begin{CJK*}{UTF8}{gbsn}
\begin{table}[!t]
\small
    \centering
    \begin{tabular}{p{0.10\linewidth} p{0.80\linewidth}}
    \toprule
        \textbf{Base.} & \textcolor{red}{myanmar} was a highly civilized country.\\ \hline
        \textbf{Tag} &  \underline{$<$start$>$ \textcolor{red}{myanmar} $<$end$>$} was a highly civilized country.\\
        \textbf{Add} & \underline{$<$start$>$ \textcolor{red}{myanmar} $<$mid$>$ \textcolor{blue}{state} $<$end$>$} was a highly civilized country.\\
        \textbf{Trans} &  \underline{$<$start$>$ \textcolor{red}{myanmar} $<$mid$>$ \textcolor{green}{缅甸} $<$end$>$} was a highly civilized country. \\
        \textbf{TransA} &  \underline{$<$start$>$\textcolor{red} {myanmar} $<$mid1$> $\textcolor{green} {缅甸}$<$mid2$> $\textcolor{blue}{state}} \underline{$<$end$>$} was a highly civilized country.\\
        \textbf{TransR} &  \underline{$<$start$>$ \textcolor{blue}{state} $<$mid$>$ \textcolor{green}{缅甸} $<$end$>$} was a highly civilized country.\\
        \textbf{HypA} & \underline{\textcolor{red}{myanmar} \textcolor{blue}{state}} was a highly civilized country.\\
    \bottomrule
    \end{tabular}
    \caption{Tagging Templates for English-Chinese \textit{source} sentences. NE (in \textcolor{red}{red}) are replaced with templates (underlined), NE hypernyms are in \textcolor{blue}{blue} and NE translations are in \textcolor{green}{green}. Best viewed in color.\label{table:tag-list}}
    \vspace{-1em}
\end{table}
\end{CJK*}

\subsection{NMT Model}

For NMT model, we used XLM introduced by \citet{conneau2020unsupervised}\footnote{https://github.com/facebookresearch/xlm}. We use the same transformer architecture as \citet{wang2019merging}: 512 embedding size, 6 encoder and decoder layer, 8 multi-attention heads. Refer to Appendix Section~\ref{sec:training} for more details. We train on both source $\rightarrow$ target and target $\rightarrow$ source direction.

\section{Experiments}
\label{sec:exp}

To evaluate our results in different resource settings, we test our methods in English-Chinese as well as English-Hausa. For English-Chinese, we randomly select 3 million pairs of sentences from MultiUN \cite{ziemski2016united} as training dataset in \textit{high}-resource setting. To evaluate English-Chinese translation, we use WMT official newstest datasets from 2017-2020. For the \textit{medium}-resource English-Hausa, we combine available parallel corpus on WMT-21 website\footnote{https://www.statmt.org/wmt21/translation-task.html} including ParaCrawl \citep{banon2020paracrawl}, Wikititles, Khamenei, and English-Hausa Opus \citep{tiedemann2012parallel}, in total of 740K parallel sentences. For simulated low-resource condition, we randomly sample 6K sentences from English-Hausa training set and use the same WA translation table in medium-resource. We evaluate English-Hausa translation on WMT official newsdev2021 and newstest2021. For all settings, We treat the WMT splits as the out-of-domain evaluations, and randomly hold out 5K valid and 5K test sentences from each training dataset as in-domain evaluation splits.

Other than evaluating translation results with multi-BLEU metric, we also investigate the accuracy of the copy mechanism. We report the copy accuracy for entity, entity translation, and hypernym whenever possible. Additionally, to understand the effect of added semantics on translating the rest of the sentence, we calculate the word translation accuracy by POS occurring before and after the tagged entity. We use 
SpaCy for English and Chinese POS tagging. With no available POS tagger for Hausa, we use alignment from FastAlign and project English POS to corresponding words in Hausa sentence, following \citet{rasooli2021wikily}.

\section{Results}
\subsection{English-Chinese (High-Resource)}
\paragraph{Tagging Improves Adequacy and Accuracy} We can see a clear improvement of around 1-4 BLEU point on average (Table~\ref{table:high-bleu}). The improvements are much larger when we evaluate it on tag-only subsets. \textbf{HypA} outperforms other methods consistently. Similar trend is observed in Chinese-English Translation (see Appx. Table~\ref{table:high-bleu-zh}).


When looking at translation accuracy (Table~\ref{table:high-copy}) of the tagged NEs, we see about 35 points improvement in translation accuracy. This is expected because copying is much easier than translating. \textbf{HypA} method, while performing better in BLEU, does not improve NE translation accuracy as much because it does not enforce "hard" copying. \textbf{Tag} method performs best in translating NEs with 91.92\% accuracy (assuming perfect word translation table). The imperfect copying result is also observed in \citet{wang2019merging} and \citet{dinu2019training}. (Error breakdown in Appx. Table ~\ref{table:copy-error-stat})

\begin{table}[t]
    \centering
    \small
    \begin{tabular}{lcc}
    \toprule
        Method              & In-Domain        & Out-of-Domain \\ \midrule
        Baseline (all)      & 33.30 $\pm$ 0.63  & 11.09 $\pm$ 0.78 \\ 
        $\boxRight$(tag-only) & 34.64 $\pm$ 2.1  & 12.21 $\pm$ 0.81 \\ \hline
        Tag (all)           & 33.77 $\pm$ 0.24 & 11.26 $\pm$ 0.91 \\ 
        $\boxRight$(tag-only)      & 36.07 $\pm$ 0.28 & 12.89 $\pm$ 1.34 \\ 
        Add (all)           & 33.69 $\pm$ 0.21 & 11.29 $\pm$ 0.81 \\ 
        $\boxRight$(tag-only)      & 35.77 $\pm$ 0.36 & 12.89 $\pm$ 1.11 \\ 
        Trans (all)         & 33.77 $\pm$ 0.04 & 11.25 $\pm$ 0.90 \\ 
        $\boxRight$(tag-only)    & 35.80 $\pm$ 0.48 & 12.97 $\pm$ 1.00 \\ 
        TransA (all)          & 33.35 $\pm$ 0.28 & 11.32 $\pm$ 0.83 \\ 
        $\boxRight$(tag-only)     & 35.37 $\pm$ 0.65 & 13.03 $\pm$ 0.98 \\ 
        TransR (all)        & 33.84 $\pm$ 0.29 & 11.18 $\pm$ 0.87 \\ 
        $\boxRight$(tag-only)   & 35.73 $\pm$ 0.61 & 12.75 $\pm$ 0.88 \\ 
        HypA (all)          & \underline{34.39} $\pm$ 0.14 & \underline{11.48} $\pm$ 0.87 \\ 
        $\boxRight$(tag-only)     & \textbf{37.54} $\pm$ 0.07 & \textbf{13.69} $\pm$ 0.95 \\ 
        
        \bottomrule
    \end{tabular}
    \caption{Mean and standard deviation of BLEU across evaluation sets for all methods in English-Chinese. Evaluation is performed on whole dataset (\textbf{all}) and on tagged sentences only (\textbf{tag-only}). Best performances are \textbf{bolded} in tag-only subsets and \underline{underscored} in all dataset. (Individual dataset results in Appx. Table~\ref{table:high-bleu-full}) \label{table:high-bleu}}
    \vspace{-1em}
\end{table}

\begin{table}[t]
    \centering
    \small
    \begin{tabular}{cccc}
    \toprule
        Method   & Entity        & Translation   & Hypernym \\ \midrule
        Baseline & -             & 55.38         & -        \\ \hline
        Tag      & 91.92         & -             & -        \\ 
        Add      & 91.02         & -             & 92.04    \\
        Trans    & \textbf{92.12}& 90.99         & -        \\
        TransA   & 91.83         & \textbf{91.27}& \textbf{92.97}    \\
        TransR   & -             & 89.12         & 91.66    \\
        HypA     & -             & 55.76         & 58.69    \\ \bottomrule
    \end{tabular}
    \caption{Mean \textit{copy} accuracy for different parts of the tag in English-Chinese across evaluation sets. We equate correct NE translation in baseline to correct translation copy. For methods containing translation, we equate translation \textit{copy} accuracy as translation accuracy. For others, we consider entity copy accuracy.
    \label{table:high-copy}}
    \vspace{-2em}
\end{table}
\vspace{-.5em}
\paragraph{Effects of Copy Mechanism on Translation}
\begin{figure*}[t]
    \centering
    \includegraphics[scale=0.61]{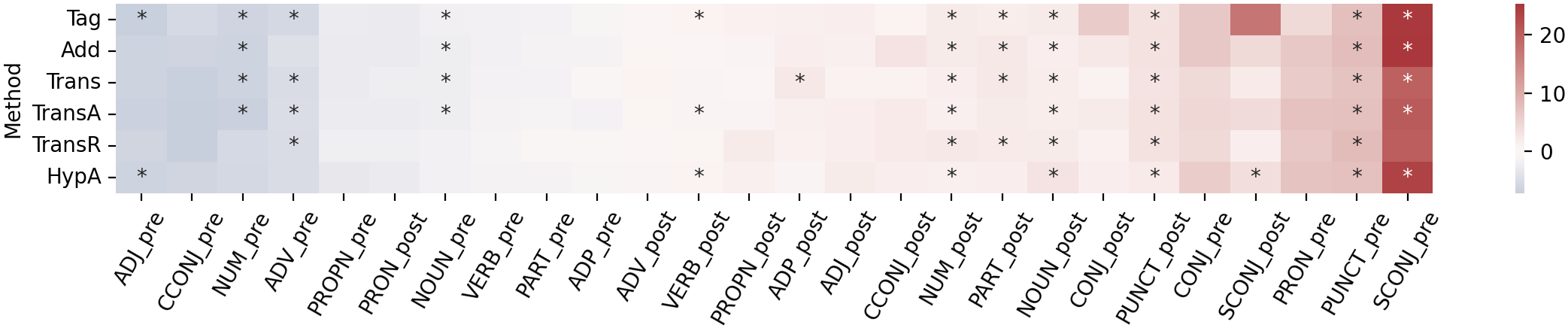}
    \caption{POS translation accuracy (percentage) difference against baseline before (\textbf{\_pre}) and after (\textbf{\_post}) the tagged entity in English-Chinese. * indicates a statistical significant difference against baseline with p-value < 0.05 \label{fig:high-pos}}
    \vspace{-1em}
\end{figure*}

As seen in Figure~\ref{fig:high-pos}, copying 
improves translation accuracy for POSs which serve as structural syntactic signals in sentences such as conjunctions, particles, punctuation while decreasing accuracy for POSs containing more semantic information that require more context to translate (verb, adjective, adverb). Qualitatively, this is equivalent to producing translations with similar sentence structures to source sentence (Appx. Table~\ref{table:qualitative-pos}). Since copying is a direct signal for models to ignore context and translate word by word for the entity, it is not surprising to see such polarizing effects on the rest of the sentences.  Unexpectedly, despite being a "soft" copy signal, \textbf{HypA} induces similar effects. We suspect that the repeating semantic of appending hypernyms after NEs yields similar signal for models to follow word-by-word order sensitive translation.

It is interesting to observe that having tag in a sentence does not uniformly improve POS given relative position to the tag. For instance, within tagged sentences, our model translates adjectives before the tag worse while translates adjectives after the tag better (than baseline). This is likely due to the adjective-noun order in English, where adjectives before tagged entities are translated worse because tagging reduces the semantic information of the entity, but adjectives after the tag are most likely not describing the entity. However, we do see consistent improvements in some categories (subjunctive conjunction, punctuation, conjunction), indicating that some POSs benefit regardless of their relative position to the tagged entity.

In Table~\ref{table:high-bleu}, we do not see significant BLEU improvement of tagging methods that contain hypernym (\textbf{Add}, \textbf{TransA}, \textbf{TransR}) over those that do not (\textbf{Tag}, \textbf{Trans}). We believe, by the same rationale above, the copy mechanism encourages models to copy, rather than using semantics of the hypernym.

\subsection{English-Hausa (Medium-Resource)}

Full English-Hausa yields similar results as English-Chinese, except that the improvements in BLEU from tagged models over baseline become marginal (Appx. Table~\ref{table:med-bleu}). 
Additionally, copy accuracy decreases from 90\% to 80\%, but remains 20\% higher than baseline (Appx. Table~\ref{table:med-copy}). 

\subsection{6K English-Hausa (Low-Resource)}
In low-resource setting, tagging does not improve BLEU, and the NE copy accuracy drops below baseline (Table~\ref{table:low-bleu}, Table~\ref{table:low-copy}). Interestingly, hypernyms are more consistently copied. We believe this is due to hypernyms having higher term frequency in the training. Compared to baseline, only \textbf{HypA} method is able to improve NE translation accuracy and obtain higher BLEU for tag-only subsets in-domain (Table~\ref{table:low-bleu}). Despite not having as high of hypernym copy accuracy, the model uses hypernym as context to improve NE translation.
\begin{table}[!h]
    \centering
    \small
    \begin{tabular}{lcc}
    \toprule
         Method      & In-Domain  & Out-of-Domain \\ \midrule
         Baseline    & \underline{7.61} $\pm$ 0.21 & 3.80 $\pm$ 3.37 \\ 
        $\boxRight$(tag-only) & 7.21 $\pm$ 0.85 & 3.40 $\pm$ 2.87 \\ \hline
        Tag (all)    & 7.39 $\pm$ 0.14 & 3.67 $\pm$ 3.12 \\ 
        $\boxRight$(tag-only) & 6.69 $\pm$ 0.79 & 3.39 $\pm$ 3.13 \\ 
        Trans (all)  & 7.45 $\pm$ 0.08 & \underline{3.91} $\pm$ 3.44 \\ 
        $\boxRight$(tag-only) & 6.99 $\pm$ 0.92 & \textbf{3.60} $\pm$ 3.44 \\ 
        HypA (all)   & 7.53 $\pm$ 0.25 & 3.52 $\pm$ 2.88 \\ 
        $\boxRight$(tag-only) & \textbf{7.82} $\pm$ 1.40 & 2.55 $\pm$ 1.89 \\ 
        
    \bottomrule
    \end{tabular}
    \caption{BLEU scores for 6K English-Hausa data. Only top performing methods are included. \label{table:low-bleu}}
    \vspace{-1em}
\end{table}

\begin{table}[ht]
    \centering
    \small
    \begin{tabular}{cccc}
    \toprule
        Method   & entity        & translation       & hypernym \\ \midrule
        Baseline & -             & 42.44             & -        \\ \hline
        Tag      & 30.72$\downarrow$& -                 & -        \\ 
        Add      & 34.48$\downarrow$& -                 & 55.66    \\
        Trans    & 37.81         & 35.69$\downarrow$ & -        \\
        TransA   & \textbf{39.01}& 37.53             & \textbf{55.91}    \\
        TransR   & -             & 30.61$\downarrow$ & 55.39    \\
        HypA     & -             & \textbf{44.77}$\uparrow$& 48.32   \\ \bottomrule
    \end{tabular}
    \caption{Mean \textit{copy} accuracy in 6K English-Hausa dataset models across evaluation sets. Arrows indicate statistical difference from baseline with p-value < 0.05. \label{table:low-copy}}
    \vspace{-1em}
\end{table}

\section{Discussion}
\vspace{-.5em}
\paragraph{Copy mechanism in low-resource.}
Copy mechanism through explicit tagging can increase NE translation accuracy in both high and medium-resource but not in low-resource condition. Learning to copy requires significant amount of data. Once tags are recognized, the semantics of the content within are ignored. Translations become structurally similar to source sentence, while focusing less on semantics. 
Without enough data, "softer" methods of augmentation (\textbf{HypA} or extra embedding \citep{moussallem2019augmenting}) that incorporates hypernym in translation is a better choice.  
Our high to low-resource results can be extended to low-low resource pairs by using high resource languages as pivots for entity linking 
(\citealp{utiyama2007comparison}, \citealp{cohn2007machine}, \citealp{wijaya-etal-2017-learning})
\vspace{-.5em}
\paragraph{Low-Resource translation affected by term frequency.}
As suggested by Table~\ref{table:low-copy}, before copy mechanism generalizes, models are more likely to copy words that occur more frequently (hypernyms). This points to potential directions in low-resource NLP in using hypernyms to bootstrap performance of other words or sentences, through data augmentation or template translation. 
\vspace{-.5em}
\paragraph{Effects of EL and WA accuracies.}
The focus of this work is to investigate copying in NMT in low resource settings. In the absence of gold labels, we assume accurate entity linking and alignment. Effects of EL and WA, though important, are orthogonal to copying, and the data requirement exists regardless of their performance. To control for their effects, we vary only the data resource setting and use the same entity linking system and word translation table in different data resource settings. Hence, data size is the only factor for BLEU and accuracy differences.
 



\section{Conclusion}
In our paper, we analyzed the tag-and-copy mechanism under different resource conditions. We found that learning to copy requires significant amount of resource often not achievable in low-resource languages. Additionally, we found that copying can induce polarizing effects on translating different POSs. It discouraged models from using contextual information, but provided "structural supervision". In low-resource setting, we found correlation between term frequency and copying accuracy. Our proposed method of appending hypernym after NEs was able to encourage slightly better translation in both low and high-resource setting.

\section{Limitation}
We want to stress that our results are negative in nature. Naive template machine translation does \textbf{not} work in the low resource regime. However, the reason behind why naive data-based template methods fail is interesting. Since entities within templates are copied word by word, the semantics of the words became insignificant to the translation. Hence, providing the hypernym or not does not make a lot of difference. 

The side effect of copying that sentences are translated with more fidelity to the original sentence structure could have interesting implications in which we design training loss objectives. In tasks like translation, where structural fidelity could be important, we could encourage such behavior by simply inserting small rule-based perturbations (always adding an extra period after every period in both source and target side, which can be removed with post-processing). Such fidelity-encouraging modification, has also been found to encourage model to generalize compositionally (glossing task) \cite{kim2021compositional}. However, when translating between languages with different word orders, order-preserving objectives such as ours would likely hurt performance. Perhaps letting target sentence be an arbitrary ordering of source sequence may provide a more abstract version of the faithfulness

Lastly, it is interesting to see hypernyms being copied more frequently than entities. Statistically, more frequent words in the training corpus do have a higher chance to be generated simply by having a higher prior. In our experiments we tried increasing the posterior by using hypernym, although there are more ways to increase the probability of the whole sentence instead using language models. If the correspondence between a complex sentence to a simpler version of the sentence can be established and fed to the model as input, perhaps it would be easier for the model to understand and translate.

\section*{Acknowledgements}
We thank Boston University for providing all the computing resources. We appreciate all of the helpful comments and feedbacks from Najoung Kim and all other anonymous reviewers.

\bibliography{custom}
\bibliographystyle{acl_natbib}

\appendix

\section{Appendix}
\label{sec:appendix}

\subsection{Text Preprocessing}
We follow default preprocessing steps in XLM repo. For English and Hausa, we use Moses \textit{tokenizer.perl} script, after which we lower-case letters and remove accents. For Chinese, we use Moses \textit{tokenizer\_PTB.perl} script. We chose to follow XLM model due to their superior performance in MT, especially in low resource settings. XLM-R was also pretrained on 100+ languages, including Hausa, making it an ideal baseline to build upon without changing variales in experiements such as vocab, objectives, model sizes, etc.

\subsection{Special Tags in XLM Model}
During tagging, in order to prevent creating additional vocabulary, we use four of the special tokens (i.e. $<$special2$>$, $<$special3$>$, $<$special4$>$, $<$special5$>$), that already exist in pretrained XLM-R model vocab, instead of actual $<$start$>$, $<$end$>$, $<$mid1$>$, and $<$mid2$>$.

\subsection{Tagging Statistics}
\begin{table}[!ht]
    \centering
    \begin{tabular}{ccc}
    \toprule
        Language Pair     & Train Size & Tag Size  \\ \midrule
        English-Hausa     & 6 K        & 1.5 K (25.6\%)    \\ 
        English-Hausa     & 746 K      & 191 K (25.6\%)    \\ 
        English-Chinese   & 2,990 K    & 816 K (27.3\%)    \\ \bottomrule
    \end{tabular}
    \caption{Tagging Statistics in Training Sets \label{table:tag-stat}}
\end{table}

\subsection{Entity Linking}
During experimentation, we have also tried more recent Entity linking systems such as BLINK \citep{li2020efficient} \footnote{https://github.com/facebookresearch/BLINK}. In reality, we find BLINK tagging less entities as well as taking a longer time. We presume this is because BLINK expects normally-cased sentences while our entity linking occurs after input sentences are lower-cased.

\subsection{Alignment Statistics}
\begin{table}[!ht]
    \centering
    \begin{tabular}{ccc}
    \toprule
        Language Pair     & Perplexity & Vocab \\ \midrule
        English-Hausa     & 64.75      & 10287 \\ 
        English-Chinese   & 131.07     & 22678 \\ \bottomrule
    \end{tabular}
    \caption{Alignment (FastAlign) Statistics in Training Sets \label{table:align-stat}}
\end{table}

\subsection{Model Training Details}
\label{sec:training}
In all of our experiments, we use the pretrained XLM-R BPE vocab with 200,000 tokens, trained on 100 lanugages \footnote{See https://github.com/facebookresearch/XLM\#the-17-and-100-languages for language details}. We use Adam optimizer, learning rate 0.0001, epoch size 300000, dropout rate of 0.1. We fix number of tokens in a batch to be around 2000. To increase batch size with GPU memory constraint, we use gradient accumulation for every four batches to increase effective batch size. For low-resource condition with 6K training sentences (see Section~\ref{sec:exp}), we change epoch size to 120,000, dropout of 0.2, and enforce minimum sentence length to 10 words. For all models, we train with translation objective only, not using any other objectives (de-noising autoencoder, online back-translation, language model objectives). All models are trained on NVIDIA V100 GPUs. Each English-Chinese model takes about 5 days to train (1 GPU time). Each English-Hausa model takes about 3 days and each English-Hausa 6K model takes about 15 hours.
\subsection{English-Chinese Full Results}
\begin{table*}[t]
    \centering
    \begin{tabular}{cccccccccc}
    \toprule
        Method   & subset   & valid & test & nd2017 & nt2017 & nt2018 & nt2019 & nt2020 & ntB2020 \\ \midrule
        Baseline & all      & 32.85 & 33.75 & 11.23 & 10.77 & 11.02 & 10.20 & 12.54 & 10.78 \\ 
        Baseline & tag-only & 33.15 & 36.12 & 13.22 & 12.69 & 12.13 & 11.30 & 12.69 & 11.20 \\ \midrule
        Tag      & all      & 33.59 & 33.94 & 11.20 & 11.38 & \underline{11.34} & 10.14 & 12.85 & 10.66 \\ 
        Tag      & tag-only & 35.86 & 36.27 & 13.72 & 14.20 & 13.18 & 11.16 & 13.85 & 11.25 \\ 
        Add      & all      & 33.53 & 33.84 & 11.15 & \underline{11.58} & 11.19 & 10.36 & 12.71 & 10.72 \\ 
        Add      & tag-only & 35.51 & 36.03 & 13.25 & 14.48 & 12.88 & 12.17 & 13.33 & 11.20 \\ 
        Trans    & all      & 33.74 & 33.80 & 11.23 & 11.10 & 10.72 & \underline{10.73} & 13.04 & 10.68 \\ 
        Trans    & tag-only & 35.45 & 36.14 & 13.46 & 13.97 & 12.40 & 12.34 & \textbf{14.04} & 11.59 \\ 
        TransA   & all      & 33.14 & 33.55 & 11.10 & 11.33 & 11.28 & 10.47 & 12.89 & 10.85 \\ 
        TransA   & tag-only & 34.90 & 35.83 & 13.50 & 13.72 & \textbf{13.54} & 12.02 & 13.84 & 11.53 \\
        TransR   & all      & 33.63 & 34.05 & 11.10 & 11.08 & 11.18 & 10.31 & 12.82 & 10.61 \\ 
        TransR   & tag-only & 35.29 & 36.16 & 13.32 & 13.65 & 12.63 & 11.85 & 13.46 & 11.56 \\  
        HypA     & all      & \underline{34.29} & \underline{34.39} & \underline{11.31} & 11.51 & 11.17 & \underline{10.73} & \underline{13.18} & \underline{10.99} \\ 
        HypA     & tag-only & \textbf{37.49} & \textbf{37.59} & \textbf{14.67} & \textbf{14.73} & 13.49 & \textbf{13.28} & 13.76 & \textbf{12.18} \\ 
        \bottomrule
    \end{tabular}
    \caption{BLEU scores across evaluation sets for all tagging methods in English-Chinese. Evaluation is performed on whole dataset and on tagged sentences only. Best performances in tagged subset are in bold. Best performances in all datasets are underscored. Each point represents a single data point. (nd2017=newsdev2017, nt2017=newstest2017, etc) \label{table:high-bleu-full}}
\end{table*}

\subsection{Chinese-English Translation Results}
\begin{table*}[t]
    \centering
    \begin{tabular}{cccccccccc}
    \toprule
        Method   & subset   & valid & test & nd2017 & nt2017 & nt2018 & nt2019 & nt2020 & ntB2020 \\ \midrule
        Baseline & all & 38.46 & 42.33 & 12.06 & 12.74 & 13 & 10.37 & 12.13 & 11.65 \\ 
        Baseline & tag-only & 43.28 & 44.87 & 13.01 & 13.81 & 14.16 & 11 & 12.88 & 12.47 \\ \midrule
        Tag      & all      & 41.47 & 42.56 & 12.53 & 12.76 & 13.06 & 10.55 & 12.48 & \underline{11.84} \\ 
        Tag      & tag-only & 44.01 & 45.13 & 14.51 & 13.87 & 14.57 & 11.94 & 13.43 & 13.17 \\ 
        Add      & all      & 41.42 & 42.37 & 12.76 & 13.14 & 12.74 & 10.38 & 12.46 & 11.83 \\ 
        Add      & tag-only & 43.82 & 44.86 & 14.73 & 14.11 & 14.33 & 11.54 & \textbf{13.67} & \textbf{13.26} \\ 
        Trans    & all      & 41.31 & 42.42 & 12.35 & 13 & 13.17 & 10.42 & 12.21 & 11.61 \\ 
        Trans    & tag-only & 43.4 & 44.8 & 13.84 & 14.26 & 14.96 & 12.14 & 13.26 & 13.02 \\ 
        TransA   & all      & 41.1 & 42.17 & 12.76 & \underline{13.21} & 13.13 & 10.66 & 12.07 & 11.52 \\ 
        TransA   & tag-only & 42.99 & 44.39 & 14.3 & \textbf{14.69} & 14.84 & 12.24 & 13.12 & 12.72 \\ 
        TransR   & all      & 41.21 & 42.28 & \underline{12.8} & 13.03 & 12.88 & \underline{10.75} & \underline{12.52} & 11.81 \\ 
        TransR   & tag-only & 43.49 & 44.75 & \textbf{15.03} & 14.26 & 14.69 & 12.26 & 13.39 & 12.82 \\ 
        HypA     & all      & \underline{41.84} & \underline{42.99} & 12.47 & 12.98 & \underline{13.29} & 10.48 & 12.2 & 11.68 \\
        HypA     & tag-only & \textbf{45.32} & \textbf{46.08} & 14.76 & 14.55 & \textbf{15.07} & \textbf{12.62} & 13.18 & 13.23\\
        \bottomrule
    \end{tabular}
    \caption{BLEU scores across evaluation sets for all tagging methods in Chinese-English. There is a consistent 0.5-2 point improvement with tagged methods over baseline.  Each point represents a single data point. \label{table:high-bleu-zh}}
\end{table*}

\subsection{Copy Efficiency In / Out of Domain}

\begin{table}[ht]
    \centering
    \begin{tabular}{ccccc}
    \toprule
             & Valid & Test & nd2017 & nt2017  \\ \midrule
        H    & 91.98 & 90.92 & 94.88 & 97.19  \\ 
        E    & 91.84 & 90.5 & 92.79 & 91.8  \\ 
        T    & 91.91 & 90.15 & 93.17 & 93.91 \\  \midrule
             & nt2018 & nt2019 & nt2020 & ntB2020 \\ \midrule
        H    & 94.45  & 94.16  & 88.48  & 91.73 \\ 
        E    & 93.76  & 93.67  & 88.02  & 92.27 \\ 
        T    & 92.37  & 92.94  & 86.41  & 89.33 \\\bottomrule
    \end{tabular}
    \caption{Copy Accuracy of TransA model across different in and out-of-domain evaluation datasets.  Each point represents a single data point. H=Hypernym, E=Entity, T=Entity translation \label{table:high-copy-io}}
\end{table}

In English-Chinese translation results, we can observe that the copy accuracy for the tags is similar across different set regardless of the domain (Table \ref{table:high-copy-io}), which is a good sign considering the drop in BLEU across the out-of-domain datasets. This indicate copy mechanism is a valuable method in translation avenues where entity translation accuracy is more valuable than adequacy (i.e. medical, scientific domain), confirming with results in \citet{pham2018towards} and \citet{dinu2019training}. 

\subsection{English-Hausa POS Accuracy Qualitative Analysis}
\begin{CJK*}{UTF8}{gbsn}
\begin{table*}[!htbp]
    \centering
    \begin{tabular}{p{0.10\linewidth} p{0.85\linewidth}}
    \toprule
        \textbf{Label} & in the \textcolor{red}{gambia} 's interim paper , it was noted that major factors in poverty among rural women include their predominance in subsistence agriculture , where they have less access than men to mechanized technologies , and the fact that , in addition to farming , they work longer hours than men carrying out household tasks .\\
        \textbf{Baseline} & the interim document of the \textcolor{red}{gambia} indicated that rural women 's poverty was mainly due to their livelihood agriculture , which was less skilled than men ; and that they were more time spent than men to run their household than men , in addition to their work .\\ \midrule
        \textbf{Tag} &  the \underline{<special2> \textcolor{red}{gambia} <special5>} interim paper indicated that the main cause of poverty among rural women was their main livelihood agriculture , less access to mechanized technologies than men ; and that in addition to farming , they were more time-consuming than men .\\
        \textbf{Add} & the \underline{<special2> \textcolor{red}{gambia} <special3> \textcolor{blue}{country} <special5>} 's interim paper noted that the main causes of poverty among rural women were their primary work in subsistence agriculture , more than men 's access to mechanical techniques , and that they would have more time than men to take their household roles in addition to their farm .\\
        \textbf{Trans} & the \underline{<special2> \textcolor{red}{gambia} <special3> \textcolor{green}{冈比亚} <special5>} 's provisional document noted that the main causes of poverty among rural women are their primary subsistence agriculture , less than men 's access to mechanized technologies , and that in addition to their farm , they are more time than men to operate household . \\
        \textbf{TransA} & in the \underline{<special2> \textcolor{red}{gambia} <special3> \textcolor{green}{冈比亚} <special4> \textcolor{blue}{country} <special5>} 's interim paper , it was noted that major factors in poverty among rural women include their predominance in subsistence agriculture , where they have less access than men to mechanized technologies , and the fact that , in addition to farming , they work longer hours than men carrying out household tasks . \\
        \textbf{TransR} & the provisional document of the \underline{<special2> \textcolor{blue}{country} <special3> \textcolor{green}{冈比亚} <special5>} indicates that the main causes of poverty among rural women are their predominance in livelihood agriculture , less access to mechanized technologies than men , and that they are more time than men to take up their housework in addition to their agricultural work . \\
        \textbf{HypA} & the interim document of the \underline{\textcolor{red}{gambia} \textcolor{blue}{country}} indicated that the main reason for poverty among rural women was their predominant livelihood farming , less than the mechanized technique of access to men ; and that they were also taking more time than men to operate their household tasks .\\
    \bottomrule
    \end{tabular}
    \caption{Translation example before post-translation tag removal. In Chinese-English translation setting, we compare all model translation results with ground truth English sentence. In all tagging methods, models tend to produce more similar sentence structures due to similar syntactic word choices. Given fixed sentence structures, there is less emphasis on translating the rest of the words that contain more semantic variations (verbs, adjectives, adverbs, etc.). NE (in \textcolor{red}{red}) are replaced with templates (underlined), NE hypernyms are in \textcolor{blue}{blue} and NE translations are in \textcolor{green}{green}. Best viewed in color.\label{table:qualitative-pos}}
\end{table*}
\end{CJK*}

\begin{table}[t]
    \centering
    \begin{tabular}{ccccc}
    \toprule
         Method & valid & test  & nd2021 & nt2021 \\ \midrule
         Base(all)  & 32.94 & 32.89 & \underline{11.31} & 21.62  \\ 
        - (tag-only)  & 35.35 & 37.12 & 11.50 & \textbf{23.18}  \\ \midrule
        Tag(all)   & \underline{33.17} & 32.99 & 10.77  & \underline{21.84}  \\
        - (tag)   & \textbf{35.91} & 37.28 & 11.86  & 23.13 \\
        Add (all)  & 32.25 & 32.62 & 11.16 & 21.42 \\ 
        - (tag-only) & 34.58 & 36.44 & 12.07 & 22.54 \\ 
        Trans(all) & 32.27 & 32.29 & 10.85 & 21.56  \\ 
        - (tag-only) & 35.45 & 36.14 & \textbf{12.01} & 22.71  \\ 
        TransA & 32.22 & 32.3 & 10.58 & 21.38 \\ 
        - (tag-only) & 33.88 & 35.94 & 11.33 & 22.56 \\ 
        TransR & 32.65 & 32.77 & 11.18 & 21.74 \\ 
        - (tag-only) & 34.74 & 36.73 & 12.38 & 22.71 \\ 
        HypA(all)  & 33.02 & \underline{33.00} & 9.59 & 20.24  \\ 
        - (tag-only)  & 35.89 & \textbf{37.39} & 8.12 & 15.42  \\ 
        
    \bottomrule
    \end{tabular}
    \caption{BLEU scores with English-Hausa full data. Each point represents a single data point. \label{table:med-bleu}}
\end{table}

\begin{table}[ht]
    \centering
    \begin{tabular}{cccc}
    \toprule
        Method   & entity        & translation   & hypernym \\ \midrule
        Tag      & 81.93         & -             & -        \\ 
        Add      & 79.16         & -             & 79.34    \\
        Trans    & \textbf{82.10}& \textbf{81.30}& -        \\
        TransR   & -             & 80.99         & \textbf{81.86}\\
        TransA   & 80.87         & 80.23         & 80.90    \\
        HypA     & -             & 61.00         & 64.29    \\
        Baseline & -             & 59.56         & -        \\ \bottomrule
    \end{tabular}
    \caption{Copy accuracy mean with English-Hausa full data. Aggregated across all evaluation datasets. \label{table:med-copy}}
\end{table}


\subsection{English-Hausa 6K Translation Results}

\begin{table}[!h]
    \centering
    \begin{tabular}{ccccc}
    \toprule
         Method    & valid & test  & nd2021 & nt2021 \\ \midrule
        Base (all)   & \underline{7.75} & \underline{7.46}  & 1.41 & 6.18  \\ 
        - (tag-only) & 6.61 & 7.81 & \textbf{1.37} & 5.43 \\  \midrule
        Tag (all)    & 7.49 & 7.29 & 1.46 & 5.87 \\ 
        - (tag-only) & 6.13 & 7.25 & 1.18 & 5.6 \\ 
        Add (all)    & 7.59 &	7.52 & 1.38	& 6.29 \\
        - (tag-only) & 6.19	& 7.61 & 1.25 &	5.48 \\
        Trans (all)  & 7.51 & 7.39 & \underline{1.48} & 6.34 \\ 
        - (tag-only) & 6.34 & 7.64 & 1.16 & 6.03 \\ 
        TransA (all) & 7.14	& 7.12 & 1.35 & 6.13\\
        - (tag-only) & 5.86	& 7.3 &	1.19 &	5.56\\
        TransR (all) & 7.35	& 7.32 &1.4 &	\underline{6.5}\\
        - (tag-only) & 5.73	& 7.22 &1.03 &	5.74\\
        HypA (all) & 7.71 & 7.35 & 1.48 & 5.55\\
        - (tag-only) & \textbf{6.83} & \textbf{8.18} & 1.21 & 3.88 \\
        
    \bottomrule
    \end{tabular}
    \caption{BLEU scores in 6K English-Hausa data for all models across individual evaluation sets. Each point represents a single data point. nd2021=newsdev2021, nt2021=newstest2021 \label{table:low-bleu-individual}}
\end{table}

\subsection{Copy Error Analysis}
\begin{table}[!ht]
    \centering
    \begin{tabular}{cccc}
    \toprule
        resource  & correct & no tag & wrong tag  \\ \midrule
        high      & 90\%   & 10\%   & 0\% \\ 
        medium    & 80\%   & 10\%   & 10\% \\ 
        low       & 35\%   & 10\%   & 55\% \\ \bottomrule
    \end{tabular}
    \caption{Approximate copy error breakdown in different resource settings. In all settings, 10 percent of the error (relatively constant throughout resource level) come from decoder not producing the copy tag while source sentence contains the tag (\textbf{no tag}). This error is mainly due to the stochastic nature of neural machine translation models, as well as potential errors in entity linking. In the remaining 0 to 55 percent of the time depending on the resource level, the tags are produced by decoder but the wrong words are being copied over (\textbf{no tag}). This type of error increases in proportion as we decrease training size. \label{table:copy-error-stat}}
\end{table}

\end{document}